\documentclass[journal]{IEEEtran}
\ifCLASSINFOpdf
\else
\fi
\ifCLASSOPTIONcompsoc
  \usepackage[caption=false,font=normalsize,labelfont=sf,textfont=sf]{subfig}
\else
  \usepackage[caption=false,font=footnotesize]{subfig}
\fi

\ifCLASSOPTIONcaptionsoff
  \usepackage[nomarkers]{endfloat}
 \let\MYoriglatexcaption\caption
 \renewcommand{\caption}[2][\relax]{\MYoriglatexcaption[#2]{#2}}
\fi

\usepackage{epsfig}
\usepackage{graphicx, float, url}
\usepackage{amsmath}
\usepackage{amssymb}
\usepackage{amsfonts}
\usepackage{multirow}
\usepackage{booktabs}
\usepackage{cite}

\begin{document}
%
\title{Deep Generative Model for Image Inpainting with Local Binary Pattern Learning and Spatial Attention}
%
%
%

\author{Haiwei~Wu, \IEEEmembership{Student~Member,~IEEE}, Jiantao~Zhou,~\IEEEmembership{Senior~Member,~IEEE}, and~Yuanman~Li, \IEEEmembership{Student~Member,~IEEE}
	\thanks{The authors are with the State Key Laboratory of Internet of Things for Smart City, and also with the Department of Computer and Information Science, Faculty of Science and Technology, University of Macau, Macau 999078, China (\emph{Corresponding Author: Jiantao Zhou, email: jtzhou@umac.mo}).}
}

%
%

\markboth{IEEE Transactions on Multimedia}%
{Shell \MakeLowercase{\textit{et al.}}: Bare Demo of IEEEtran.cls for IEEE Journals}
%



\maketitle

\begin{abstract}
Deep learning (DL) has demonstrated its powerful capabilities in the field of image inpainting. The DL-based image inpainting approaches can produce visually plausible results, but often generate various unpleasant artifacts, especially in the boundary and highly textured regions. To tackle this challenge, in this work, we propose a new end-to-end, two-stage (coarse-to-fine) generative model through combining a local binary pattern (LBP) learning network with an actual inpainting network. Specifically, the first LBP learning network using U-Net architecture is designed to accurately predict the structural information of the missing region, which subsequently guides the second image inpainting network for better filling the missing pixels. Furthermore, an improved spatial attention mechanism is integrated in the image inpainting network, by considering the consistency not only between the known region with the generated one, but also within the generated region itself. Extensive experiments on public datasets including \texttt{CelebA-HQ}, \texttt{Places} and \texttt{Paris StreetView} demonstrate that our model generates better inpainting results than the state-of-the-art competing algorithms, both quantitatively and qualitatively. The source code and trained models will be made available at {https://github.com/HighwayWu/ImageInpainting}.
\end{abstract}


%
\IEEEpeerreviewmaketitle

\section{Introduction}
\IEEEPARstart{I}{mage} inpainting is to fill the missing region of an image with plausible contents. It has a wide range of applications in the field of computer vision, e.g., repairing damaged photos or removing unwanted objects. The major challenge faced by image inpainting is the generation of visually realistic and semantically plausible contents for the missing region that is consistent with the known part.

Several traditional approaches \cite{efros1999texture, efros2001image, depth2011, bertalmio2003simultaneous, criminisi2004region, virtual2011, sun2005image, hays2007scene, simakov2008summarizing, keyframebased2018, barnes2009patchmatch, spatio2018, holefilling2019, structureguide2018} attempted to solve the inpainting problem via the image-level texture synthesis. Sun \emph{et al.} \cite{sun2005image} proposed to adopt user-specified curves to complete missing structures, and then fill the missing region via patch-based texture synthesis. Hays and Efros \cite{hays2007scene} built a huge database of photographs, from which similar patches can be fetched for image inpainting. Similarly, Simakov \emph{et al.} \cite{simakov2008summarizing} suggested an approach based on the bidirectional patch similarity to better summarize visual data for re-targeting, object removal and image inpainting. Barnes \emph{et al.} \cite{barnes2009patchmatch} designed the Patch-Match algorithm, significantly speeding up the searching of similar patches, which could be used for image inpainting. Recently, Huang \emph{et al.} \cite{structureguide2018} presented a novel structure-guided inpainting method by maintaining the neighborhood consistence and structure coherence of inpainted regions. As these inpainting methods essentially assumed that the missing region shares the same structural features with the known one, they cannot create novel contents for the challenging cases where the missing region involves complex structures (e.g., faces) and high-level semantics \cite{yu2018generative}.

\begin{figure}[t!]
	\subfloat{
		\includegraphics[width=.226\linewidth]{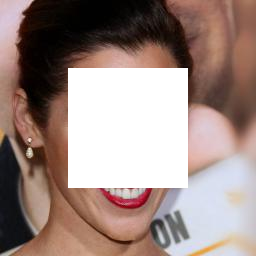}
	}
	\subfloat{
		\includegraphics[width=.226\linewidth]{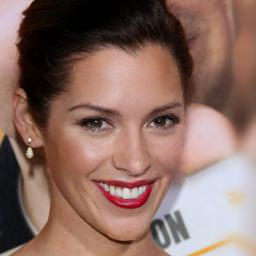}
	}
	\subfloat{
		\includegraphics[width=.226\linewidth]{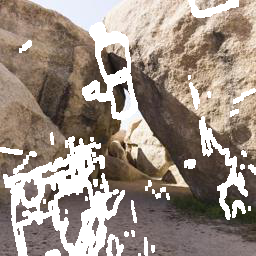}
	}
	\subfloat{
		\includegraphics[width=.226\linewidth]{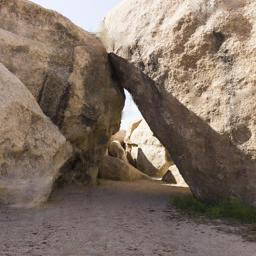}
	}
	\newline
	\subfloat{
		\includegraphics[width=.226\linewidth]{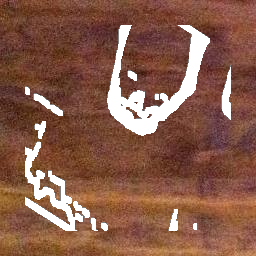}
	}
	\subfloat{
		\includegraphics[width=.226\linewidth]{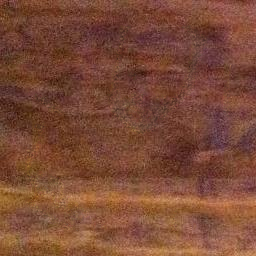}
	}
	\subfloat{
		\includegraphics[width=.226\linewidth]{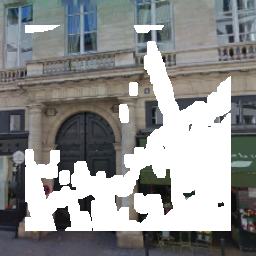}
	}
	\subfloat{
		\includegraphics[width=.226\linewidth]{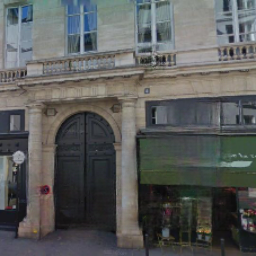}
	}
	\caption{Inpainting results generated by our proposed model on images of face, natural scene, texture and street view. In each pair, the left is the input image with centering or irregular mask, and the right is the inpainting result.}
	\label{fig:heading_figure}
\end{figure}

With the rapid development of deep convolutional neural networks (CNNs) and generative adversarial networks (GANs) \cite{goodfellow2014generative} in recent years, many deep generative models \cite{dolhansky2018eye, iizuka2017globally, isola2017image, li2019generative, liu2018image, liu2019coherent, ma2019coarse, nazeri2019edgeconnect, pathak2016context, ren2019structureflow, sagong2019pepsi, song2018contextual, song2018segmentation, wang2019musical, wang2018image, xiao2019cisinet, xie2019image, xiong2019foreground, yan2018shift, yeh2017semantic, yu2018generative, zeng2019learning, zhao2019parallel} have been proposed for image inpainting, achieving promising results. Generally, these models employ a convolutional encoder-decoder network, where the encoder extracts high-level information from image-level pixels, from which the decoder then generates semantically coherent contents. During this process, adversarial networks are often jointly trained to promote consistency between the generated and existing pixels. These deep generative models were reported to be able to produce plausible new contents, e.g., faces, objects and scenes \cite{yu2018generative}. Nevertheless, they also tend to generate unpleasant boundary artifacts, distorted structures, and blurry textures that are incoherent with the available parts. As pointed out in \cite{yu2018generative}, such inferior performance may be due to the ineffectiveness of CNNs in modeling long-term correlations between the distant contextual information and the missing region.

In this paper, inspired by the image inpainting procedure conducted by human artists, we propose a new end-to-end, coarse-to-fine deep generative image inpainting model that consists of two networks. The first network, called Local Binary Pattern (LBP) learning network, aims to recover the LBP feature of the missing region, based on the known region. One of the reasons why we select LBP feature under this circumstance is that it contains a great amount of structural information, capable of well guiding the subsequent image inpainting task. As verified in \cite{waller2013image}, an image visually close to the original one could be reconstructed solely from its LBP feature. Also, from the perspective of practical implementation, LBP is easy to be computed and very few parameters are involved. The second network, called image inpainting network, performs the actual operations for filling the missing pixels, by using the learned LBP feature as the guidance. To further promote the semantic relevance of the filled pixels, we propose to integrate a novel spatial attention layer into the image inpainting network. The proposed attention layer has a unique mechanism that models the correlations not only between the known region with the generated one, but also within the generated region itself. The latter correlation was largely ignored by \emph{all} the existing methods \cite{liu2019coherent, sagong2019pepsi, song2018contextual, wang2019musical, yan2018shift, yu2018generative, zeng2019learning}. As a result, our proposed generative model leads to better global and local consistency in the inpainting results. Meanwhile, in order to make the training process more stable, we design a multi-level loss such that multi-level features could be optimized. Experiments on three publicly available datasets \texttt{CelebA-HQ} \cite{karras2017progressive}, \texttt{Places} \cite{zhou2017places} and \texttt{Paris StreetView} \cite{doersch2012makes} demonstrate that our proposed model can generate better results than the state-of-the-art competitors, both quantitatively and qualitatively. Fig. \ref{fig:heading_figure} shows some example results. We also would like to emphasize that, among all the two-stage networks for the image inpainting \cite{liu2019coherent, nazeri2019edgeconnect, ren2019structureflow, song2018contextual, xiong2019foreground, yu2018generative}, the key differences lie in how to design these two networks, and naturally, different designs could lead to dramatically different inpainting results.

Our major contributions can be summarized as follows:
\begin{itemize}
	\item We propose a new end-to-end, coarse-to-fine deep generative model that incorporates LBP learning to provide structural information for the inpainting task. A multi-level loss is also designed to ensure more stable training process.
	\item We introduce a novel spatial attention layer that models the correlations not only between the known region and the filled one, but also within the filled region itself. This leads to better global and local consistency of the inpainting results.
	\item Our model achieves better inpainting performance in comparison with several state-of-the-art methods \cite{nazeri2019edgeconnect, yan2018shift, yu2018generative, yu2019free} over a variety of challenging datasets including \texttt{CelebA-HQ}, \texttt{Places} and \texttt{Paris StreetView}.
\end{itemize}

The rest of this paper is organized as follows. Section \ref{sec:related work} reviews the related works on LBP and deep generative models for image inpainting. Section \ref{sec:methods} presents our proposed model. Experimental results are given in Section \ref{sec:experiments} and Section \ref{sec:conclusions} concludes.

\section{Related Works}\label{sec:related work}
\subsection{Local Binary Pattern (LBP)}
\begin{figure}[t]
	\centering
	\includegraphics[width=0.45\textwidth]{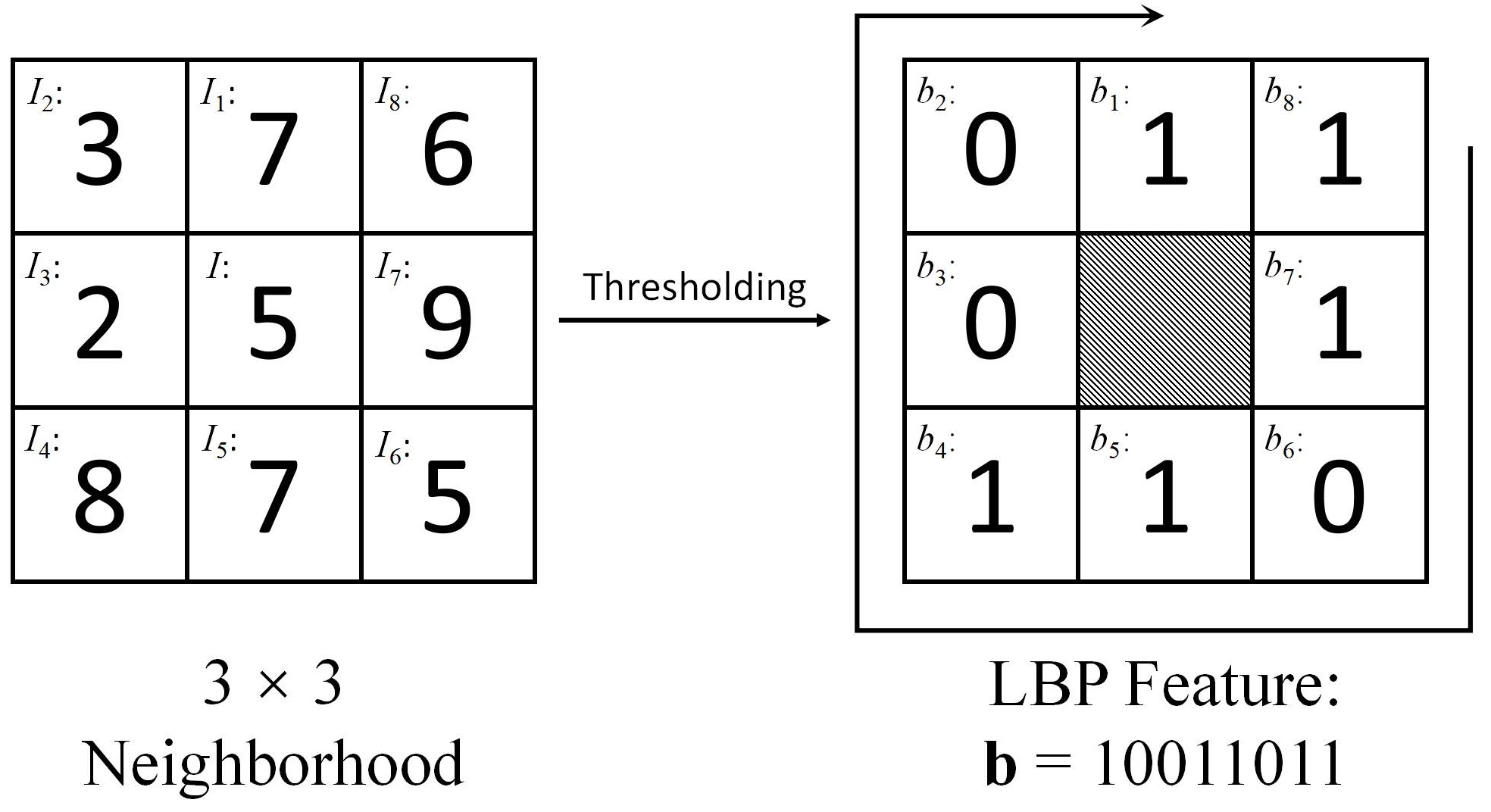}
	\caption{An example of the LBP extraction. Left is the original $3 \times 3$ neighborhood. Right is the thresholded neighborhood, and the LBP feature of the centering pixel $I$ is $\mathbf{b} = 10011011$. }
	\label{fig:LBP_extraction}
\end{figure}

LBP is a simple yet very effective texture descriptor originally proposed by Ojala \emph{et al.} \cite{ojala1996comparative}. The LBP feature extraction process is to label each pixel of an image by thresholding its spatial neighborhood. Specifically, to extract the LBP feature associated with the pixel $I$, we first obtain its $3\times 3$ neighborhood denoted by $I_1, I_2, \cdots, I_{8}$. Then the LBP feature associated with $I$ is an 8-bit long binary string $\mathbf{b} = b_1, b_2, \cdots, b_{8}$, where

\begin{equation}
b_i=
\begin{cases}
0& \mathrm{if}~ I_i \leq I\\
1& \mathrm{otherwise}
\end{cases},
\mathrm{for}~i = 1, 2,\cdots, 8.
\end{equation} An example of the LBP feature extraction is illustrated in Fig. \ref{fig:LBP_extraction}.

LBP feature essentially records the relative ordering within a block of pixels, capturing the information of edges, spots and other local structures \cite{zhang2010local}. LBP shows very good performance in many vision tasks, e.g., unsupervised texture segmentation \cite{ojala1999unsupervised}, face recognition \cite{ahonen2006face} and image reconstruction \cite{waller2013image}.

\subsection{Image Inpainting by Deep Generative Models}

Many DL- and GAN-based inpainting methods have been proposed in recent years, achieving promising results. Pathak \emph{et al.} \cite{pathak2016context} pioneered the research in this direction by training deep generative adversarial networks for inpainting large holes in images. However, the proposed networks cannot satisfactorily maintain global consistency and tends to produce severe visual artifacts. Iizuka \emph{et al.} \cite{iizuka2017globally} designed a generative network with two context discriminators to encourage global and local consistency, where the global discriminator evaluates whether the image is coherent as a whole, and the local discriminator ensures local consistency of the generated patches. Instead of merely using the features of the encoder layer, Yan \emph{et al.} \cite{yan2018shift} proposed an attention mechanism, which jointly uses the encoder layer and the corresponding decoder layer to estimate the missing features. To further improve the attention mechanism, Wang \emph{et al.} \cite{wang2019musical} suggested a multi-stage image contextual attention learning strategy to deal with the rich background information flexibly while avoiding abuse them. Meanwhile, several works \cite{liu2018image, yu2019free} adopted partial or gated convolutions to reduce the color discrepancy and blurriness, where the convolutions are masked, re-normalized, and operated only on the known region.

\begin{figure*}[t]
	\centering
	\includegraphics[width=\textwidth]{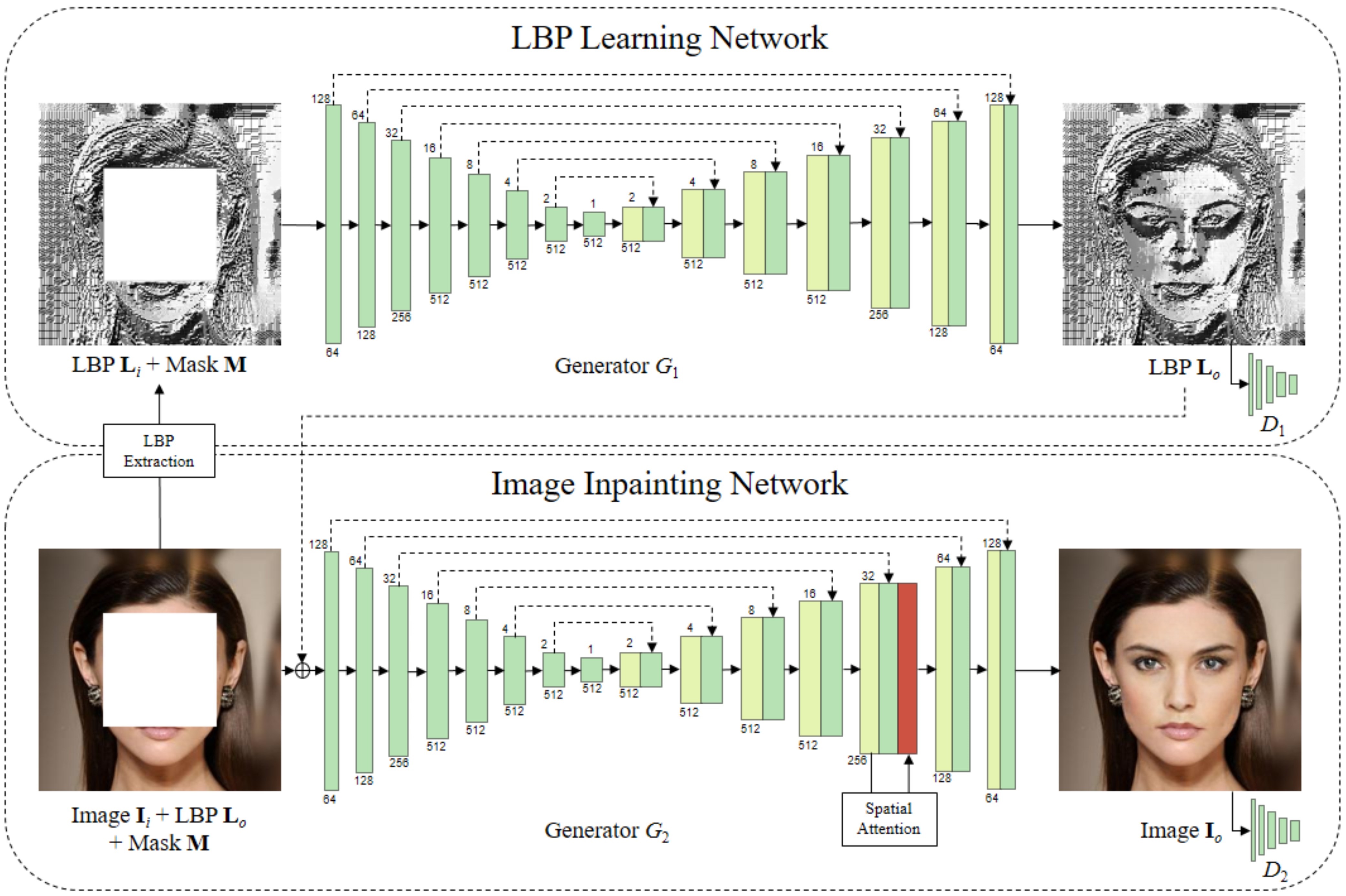}
	\caption{Overview of our proposed generative inpainting network. Note that a spatial attention layer is concatenated in the decoder of the image inpainting network. The number above each layer represents the size of the resolution, while the number below means the dimension.}
	\label{fig:framework}
\end{figure*}

Attempting to further improve the inpainting performance, there is a recent trend of using two-stage networks, where the first stage estimates the missing structures and the second stage aims to generate the final results assisted by the estimated structural information. Along this line, Yu \emph{et al.} \cite{yu2018generative} proposed to use a simple dilated convolutional network in the first stage to rough out the missing contents, and then integrated the contextual attention in the second stage. Song \emph{et al.} \cite{song2018contextual} introduced a patch-swap layer to propagate the high-frequency texture details from the boundary to the hole, where a VGG network is used as the feature extractor. By incorporating the prior knowledge on local patches continuity, Liu \emph{et al.} \cite{liu2019coherent} suggested a coherent semantic attention layer to model the semantic relevance between the holes, and iteratively optimize them to achieve better spatial consistency. Ren \emph{et al.} \cite{ren2019structureflow} designed the StructureFlow, which employs edge-preserved smooth images to train a structure reconstructor in the first stage, and then uses a texture generator with appearance flow in the second stage to yield the image details. Xiong \emph{et al.} \cite{xiong2019foreground} built a foreground-aware image inpainting model that detects and completes the foreground contour first, and then fills the missing region using the predicted contour as a guidance. Furthermore, Nazeri \emph{et al.} \cite{nazeri2019edgeconnect} proposed an edge-connect model that comprises of an edge generator followed by an image completion network. These two-stage methods demonstrate promising visual results by using the learned information to assist the ultimate inpainting task.

\section{Proposed Deep Generative Model with LBP Learning and Spatial Attention}\label{sec:methods}

\begin{figure}[t]
	\centering
	\includegraphics[width=.48\textwidth]{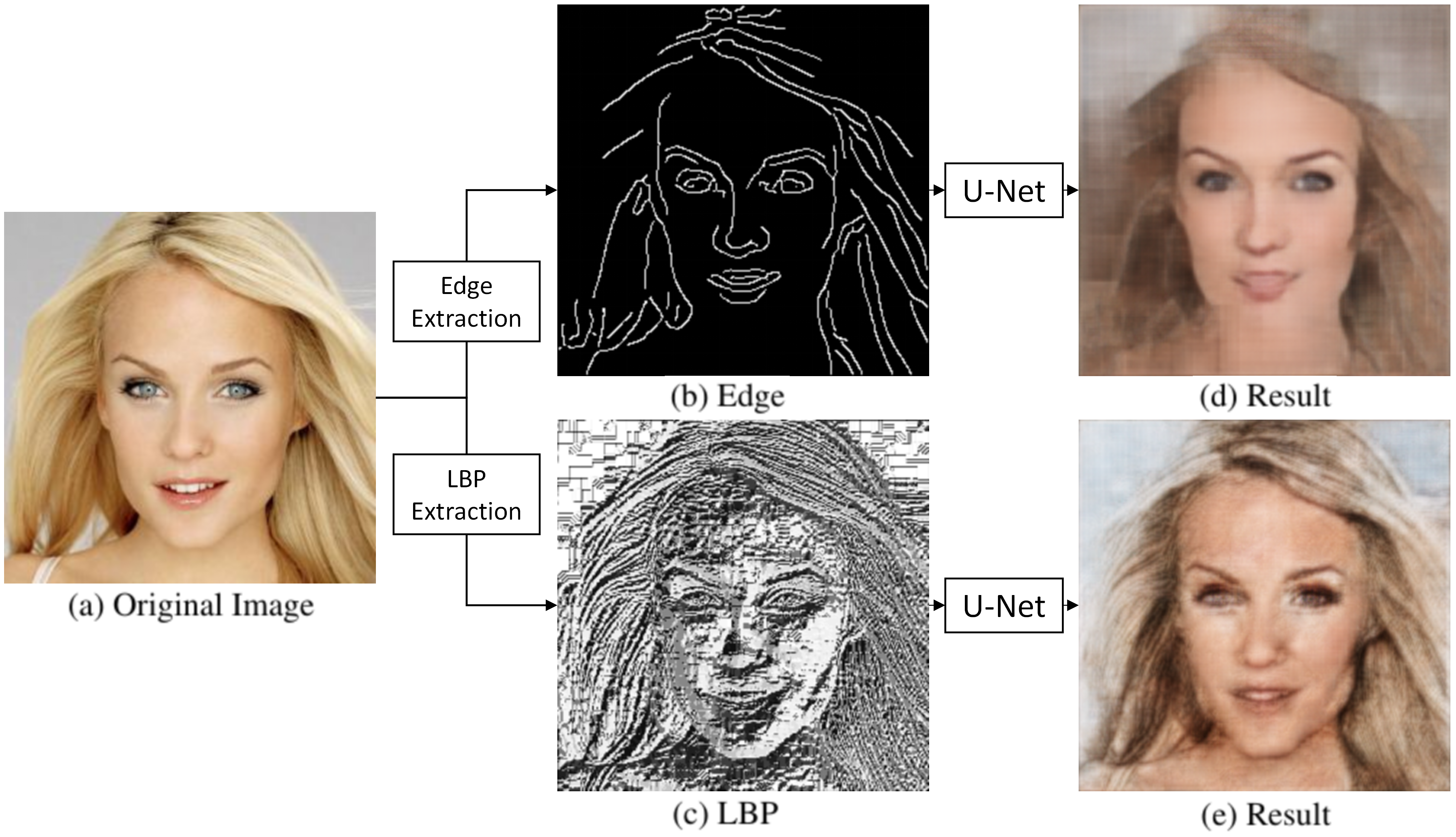}
	\caption{Image reconstruction results from edges and LBP feature. Here, Canny edge detection \cite{canny1986computational} is used with $\sigma = 2$ (the recommended setting in \cite{nazeri2019edgeconnect}). }
	\label{fig:Edge_LBP_Unet}
\end{figure}

The proposed deep generative model for image inpainting falls into the category of two-stage scheme, which can be shown in Fig. \ref{fig:framework}. As mentioned previously, different designs of these two stages could lead to dramatically different inpainting performance. Our model consists of two networks: LBP learning network and image inpainting network. The first network predicts the LBP information of the missing region, serving as a guidance for the actual image inpainting task in the second network. The major reason why we choose the LBP in the first network is because LBP contains richer amount of structural information, compared with other alternatives, e.g., edges. An inspiring example is given in Fig. \ref{fig:Edge_LBP_Unet}, where we compare the quality of the reconstructed images from the edges and the LBP. As can be seen, the reconstructed result from LBP is much better than the one obtained from the edges, especially in the fine textured regions, e.g., the hairs. In addition, LBP feature extraction is of low complexity, and involves very few parameters. In contrast, some other structural information (e.g., edges and contours) extractions typically involve many parameters, e.g., the pre-filtering strength and the threshold for the edge response, whose optimal setting should vary for different images. These facts would suggest that LBP is a more appropriate candidate for providing structural information to be incorporated into the image inpainting network.

Both networks follow an adversarial model \cite{goodfellow2014generative}. More specifically, each network contains a generator based on U-Net architecture \cite{ronneberger2015u}, and a discriminator based on the PatchGAN \cite{isola2017image}. Let $\mathbf{I}_i$ be an input image with white pixels filled in the missing region, and $\mathbf{M}$ be the corresponding binary mask, where 1's are assigned to the known region and 0's elsewhere. $\mathbf{M}$ is randomly sampled from the mask dataset and can be either centering or irregular. Denote $\mathbf{L}_i$ as the LBP extracted from $\mathbf{I}_i$ in the grayscale channel. At the training stage, the generator of the LBP learning network $G_1$ takes the pair $(\mathbf{L}_i, \mathbf{M})$ as input, and outputs the learned LBP $\mathbf{L}_o$, where the LBP feature for the missing region has been restored. During this process, the discriminator $D_1$ works together with the $G_1$ to produce the result $\mathbf{L}_o$. Upon having a well-estimated LBP, we then use it to guide the inpainting process in the inpainting network. Specifically, the generator $G_2$ takes ($\mathbf{I}_i$, $\mathbf{L}_o$, $\mathbf{M}$) as input, and outputs the final inpainting result $\mathbf{I}_o$, with the assistance of the discriminator $D_2$. At the testing stage, the procedure is similar, but without the need of using the two discriminators $D_1$ and $D_2$.

In the following, we present the details regarding the LBP learning network and the image inpainting network.

\subsection{LBP Learning Network}

The LBP learning network is formed by two components: the generator $G_1$ and the discriminator $D_1$. For $G_1$, we adopt a pruned U-Net architecture \cite{ronneberger2015u} composed of an encoder and a decoder. In the encoder, each layer has a $4\times4$ convolution, an Instance Norm \cite{ulyanov2016instance} and a LeakyReLU \cite{xu2015empirical} with $\alpha=0.2$. The decoder has a symmetric structure, except that the convolution and LeakyReLU are replaced with the deconvolution and ReLU \cite{nair2010rectified}, respectively. Additionally, skip connections are used to concatenate the features from each layer of the encoder with the corresponding layer of the decoder. Experimentally, we find that the dilated convolutions in the original U-Net architecture \cite{ronneberger2015u} bring negligible improvements to the final inpainting results. We hence prune the U-Net architecture by removing the dilated convolutions, so as to reduce the number of model parameters, which could speed up the training process. For the $D_1$, we adopt the PatchGAN architecture \cite{isola2017image}.

In addition to the network architecture, another important ingredient for achieving a desirable LBP learning network is the loss function. To better deal with the training instability, we design a multi-level loss to penalize the feature-domain deviation of the model. Specifically, let $\mathbf{L}_g$ be the ground-truth LBP, and its corresponding high-level features be $\Phi_h(\mathbf{L}_g)$, where $h$ is the layer index within $G_1$. As the training direction, $\Phi_h(\mathbf{L}_g)$ can optimize high-level features of the $G_1$ globally. The multi-level loss function is then defined as:

\begin{equation}\label{loss:multi-level}
\mathcal{L}_m = \sum_{h \in \mathcal{H}}  ||\Phi_h(\mathbf{L}_{o}) - \Phi_h(\mathbf{L}_{g})||_2,
\end{equation} where $\mathcal{H}$ accommodates the indexes of \emph{all} the convolution and deconvolution layers in $G_1$.

Besides the multi-level loss, the reconstruction loss and the adversarial loss need to be included as well. In this work, the reconstruction loss is naturally defined as:

\begin{equation}\label{loss:recon}
\mathcal{L}_{r} = ||\mathbf{L}_{o} - \mathbf{L}_{g}||_2.
\end{equation} Also, the adversarial loss \cite{yan2018shift} can be calculated as follows:

\begin{equation}\label{loss:adv}
\begin{aligned}
\mathcal{L}_{a} = \min\limits_{G_1}\max\limits_{D_1} &\mathbb{E}_{\mathbf{L}_{g}}[\mbox{log}D_1(\mathbf{L}_{g})] +\\ &\mathbb{E}_{\mathbf{L}_{i}}[\mbox{log}(1-D_1(G_1(\mathbf{L}_{i}, \mathbf{M})))].
\end{aligned}
\end{equation}

Finally, the loss function for the LBP learning network is defined by integrating the above three types of loss.

\begin{equation}
\mathcal{L}_{LBP} = \lambda_m\mathcal{L}_m + \lambda_r\mathcal{L}_r + \lambda_{a}\mathcal{L}_{a},
\end{equation}
where $\lambda_m$, $\lambda_r$ and $\lambda_{a}$ are the parameters trading off different types of loss, whose settings will be clarified in the next Section.

\subsection{Image Inpainting Network}\label{sec:image_inpainting}

The architecture of the image inpainting network is similar to our LBP learning network, except that $\mathbf{I}_{i}$, $\mathbf{L}_{o}$ and $\mathbf{M}$ are used as inputs together, and a newly designed spatial attention layer is embedded in the fifth layer of the decoder. The feature map of the spatial attention layer is of size $32 \times 32$, aiming at more effectively modeling the correlations not only between the known region with the filled one, but also within the filled region itself. In the following, let us first explain the loss function for the image inpainting network, and then present the details of our proposed spatial attention layer.

\subsubsection{Loss Function for the Inpainting Network}

To better optimize the high-level features of the image inpainting network, we further introduce two loss terms, namely, the perceptual loss \cite{johnson2016perceptual} and the style loss \cite{gatys2016image}. Specifically, the perceptual loss penalizes the inpainting results that are not perceptually similar to the ground-truth $\mathbf{I}_g$, and it can be defined as:
\begin{equation}
\mathcal{L}_p=\sum_{h\in \mathcal{A}}||\varphi_h(\mathbf{I}_{o})-\varphi_h(\mathbf{I}_{g})||_2,
\end{equation}
where $\varphi_h$ is the activation map corresponding to the $h$-th layer of an ImageNet-pretrained VGG-16 network. The set $\mathcal{A}$ is formed by the layer indexes of $\rm conv2\_1$, $\rm conv3\_1$, $\rm conv4\_1$ layers. On the other hand, the style loss is used to measure the differences between the covariances of the activation maps, which is an effective strategy to eliminate the ``checkerboard'' artifacts cased by deconvolution layers \cite{sajjadi2017enhancenet}. Typically, the style loss can be defined as:
\begin{equation}
\mathcal{L}_s= \sum_{h\in \mathcal{A}}|| \mathbf{G}^{\varphi_h}(\mathbf{I}_{o}) - \mathbf{G}^{\varphi_h}(\mathbf{I}_{g}) ||_2,
\end{equation}
where $\mathbf{G}^{\varphi_h}$ is a $3 \times 3$ Gram matrix constructed from the activation map $\varphi_h$.

For the total loss function of the image inpainting network, we also add the multi-level loss, the reconstruction loss and the adversarial loss, which can be similarly defined as in (\ref{loss:multi-level}), (\ref{loss:recon}) and (\ref{loss:adv}), respectively. Finally, the loss function of the image inpainting network can be expressed as:

\begin{equation}
\mathcal{L}_{Img} = \lambda_m\mathcal{L}_m + \lambda_r\mathcal{L}_r + \lambda_{a}\mathcal{L}_{a} + \lambda_p\mathcal{L}_p + \lambda_s\mathcal{L}_s,
\end{equation}
where $\lambda_m$, $\lambda_r$, $\lambda_{a}$, $\lambda_p$ and $\lambda_s$ are parameters used for trading off different losses. 

\subsubsection{Spatial Attention Layer}\label{sec:attention_layer}

Another crucial element in our proposed scheme is a new spatial attention layer, further improving the semantic consistency, not only between the known region and the filled region, but also within the filled region itself. This is quite different from the  existing attention models \cite{liu2019coherent, sagong2019pepsi, song2018contextual, wang2019musical, yan2018shift, yu2018generative, zeng2019learning}, which only paid attention to the relevant patches of the known region, while totally ignoring the correlations among the generated patches.

More specifically, let $\Phi_h(\mathbf{I}_i)$ be the feature map of the $h$-th layer in the generator $G_2$ when using $\mathbf{I}_i$ as the input. Denote $\Omega$ and $\bar{\Omega}$ as the missing region and the known region of $\Phi_h(\mathbf{I}_i)$, respectively. We extract all $1 \times 1$ patches $\{\mathbf{P}_j\}_{j=1}^K$ from $\Phi_h(\mathbf{I}_i)$ and group them into two sets $\mathcal{P}$ and $\bar{\mathcal{P}}$, where


\begin{eqnarray}
\mathcal{P} = \Big\{\mathbf{P}_j| \mathbf{P}_j \in  \Omega \Big\}, \\
\bar{\mathcal{P}} = \Big\{\bar{\mathbf{P}}_k| \bar{\mathbf{P}}_k \in  \bar{\Omega} \Big\}.
\end{eqnarray}
For each patch $\mathbf{P}_j \in \mathcal{P}$, its intra- cosine similarities within $\mathcal{P}$ and inter- cosine similarities with $\bar{\mathcal{P}}$ can be respectively computed as

\begin{equation}\label{eq:S}
S_{j, k} = \Big \langle \frac{\mathbf{P}_j}{||\mathbf{P}_j||}, \frac{\mathbf{P}_{k}}{||\mathbf{P}_{k}||} \Big \rangle,~~ \mathbf{P}_{k}\in \mathcal{P},
\end{equation}
\begin{equation}\label{eq:Sbar}
\bar{S}_{j, k} = \Big \langle \frac{\mathbf{P}_j}{||\mathbf{P}_j||}, \frac{\bar{\mathbf{P}}_{k}}{||\bar{\mathbf{P}}_{k}||} \Big \rangle,~~ \bar{\mathbf{P}}_{k}\in \bar{\mathcal{P}}.
\end{equation} Upon computing all $S_{j, k} $'s and $\bar{S}_{j, k}$'s, we can readily obtain the top-$T$ similar patches for $\mathbf{P}_j$ from $\mathcal{P}$ and $\bar{\mathcal{P}}$, respectively. Let $\mathcal{N} = \{n_1, ..., n_T\}$ and $\bar{\mathcal{N}} = \{\bar{n}_1, ..., \bar{n}_T\}$ record the indexes of these top-$T$ similar patches in $\Omega$ and $\bar{\Omega}$, respectively. The process of similarity search can be conducted via a convolutional layer, as explained in \cite{yan2018shift, yu2018generative}. We then propose to update each $\mathbf{P}_j \in \mathcal{P}$ via a non-local mean \cite{buades2005non} strategy:

\begin{equation}\label{eq:update-patch}
\begin{aligned}
\mathbf{P}_j^* = \sum_{k \in \mathcal{N}} \frac{\exp(S_{j, k})}{Z_j} \mathbf{P}_{k}+  \sum_{k \in \bar{\mathcal{N}}} \frac{\exp(\bar{S}_{j, k})}{Z_j} \bar{\mathbf{P}}_{k},
\end{aligned}
\end{equation} where $Z_j$ is the normalization factor:
\begin{equation}
Z_j = \sum_{k \in \mathcal{N}} \exp(S_{j, k})+  \sum_{k \in \mathcal{\bar{N}}} \exp(\bar{S}_{j, k}).
\end{equation} Therefore, the updated $\mathbf{P}_j^*$ absorbs the information from not only the top-$T$ most similar feature patches in the known region, but also from the ones in the missing region. As expected and will be verified experimentally, the new $\mathbf{P}_j^*$ could better promote semantic coherence both globally and locally.

\begin{figure}[t]
	\centering
	\includegraphics[width=0.48\textwidth]{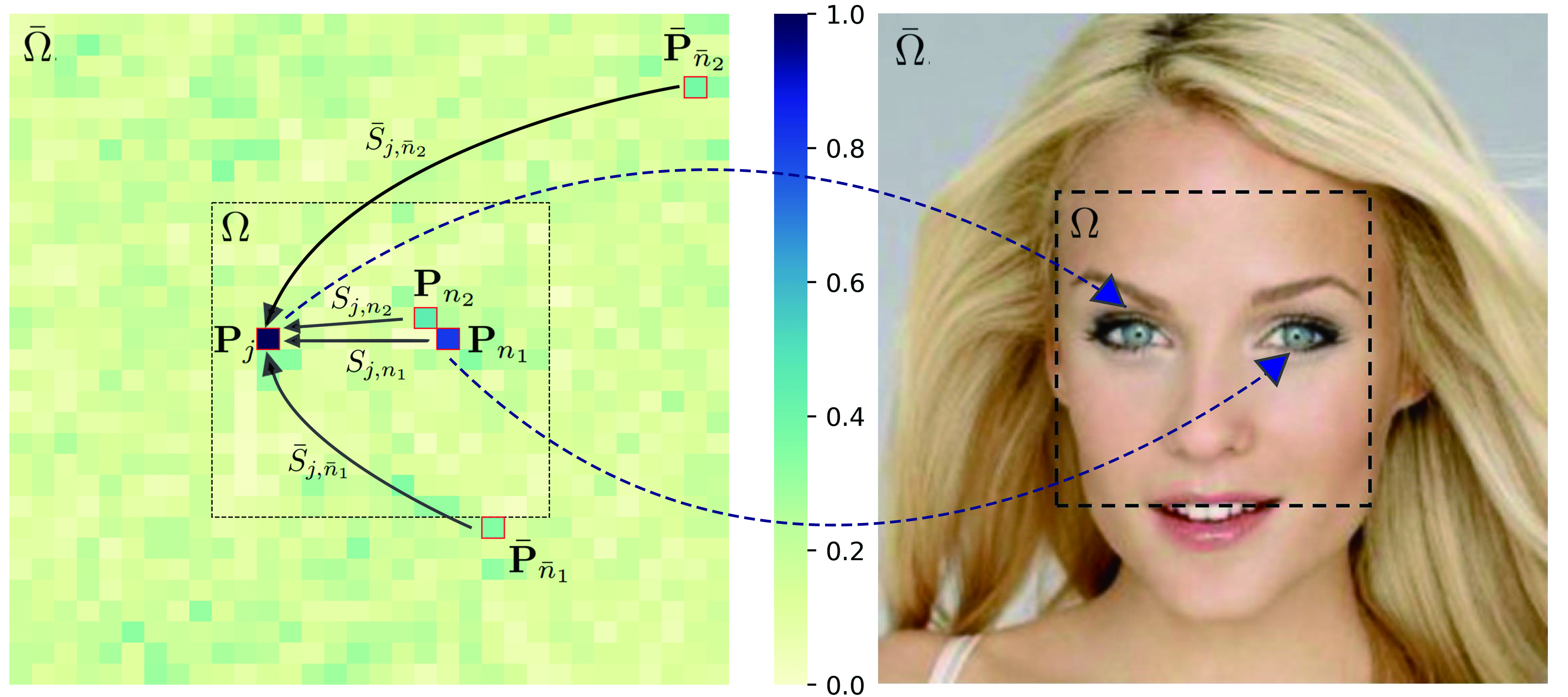}
	\caption{An example of our spatial attention layer. The depth of the color in the left image represents the level of similarity with the feature patch $\mathbf{P}_j$. The right image helps better understand the semantic correlations in the pixel domain.}
	\label{fig:illustration}
\end{figure}

An illustrative example is given in Fig. \ref{fig:illustration} where the layer index $h=13$, the number of the most similar patches $T = 2$ and $\Omega$ is a centering rectangle region indicated by dotted lines. The feature patch $\mathbf{P}_j$ in Fig. \ref{fig:illustration} would correspond to the left eye in the pixel domain. $\mathbf{P}_{n_1}$ and $\mathbf{P}_{n_2}$ are the patches having the highest similarities with $\mathbf{P}_j$ in $\Omega$, while $\bar{\mathbf{P}}_{\bar{n}_1}$ and $\bar{\mathbf{P}}_{\bar{n}_2}$ are the most similar two patches in $\bar{\Omega}$. When the missing region is included in the attention scope, we can find the most relevant patch $\mathbf{P}_{n_1}$ with $S_{j, n_1} = 0.8$. Very likely, $\mathbf{P}_{n_1}$ would correspond to the right eye in the pixel domain. However, if the search range is constrained to the known region only, then the globally most similar patch $\mathbf{P}_{n_1}$ will be missed out. Therefore, by paying extra attention to the generated patches, we may not only provide more relevant patches as references for optimization, but also enhance the network's ability to understand the semantic correlations within the missing region.

\begin{figure*}[t]
	\centering
	\subfloat{
		\centering
		\includegraphics[width=.13\textwidth]{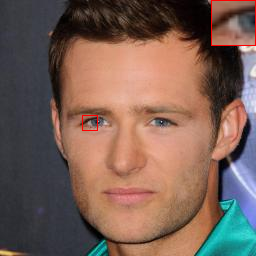}
	}
	\subfloat{
		\centering
		\includegraphics[width=.13\textwidth]{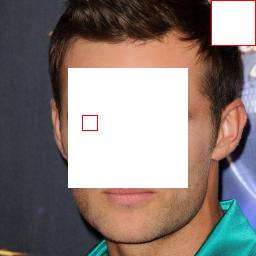}
	}
	\subfloat{
		\centering
		\includegraphics[width=.13\textwidth]{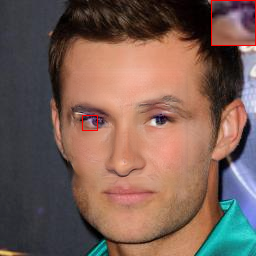}
	}
	\subfloat{
		\centering
		\includegraphics[width=.13\textwidth]{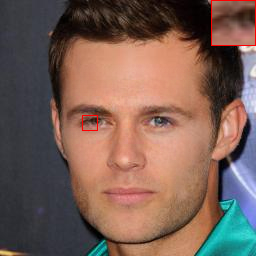}
	}
	\subfloat{
		\centering
		\includegraphics[width=.13\textwidth]{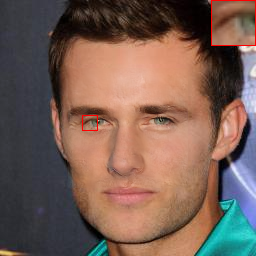}
	}
	\subfloat{
		\centering
		\includegraphics[width=.13\textwidth]{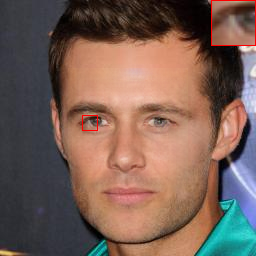}
	}
	\subfloat{
		\centering
		\includegraphics[width=.13\textwidth]{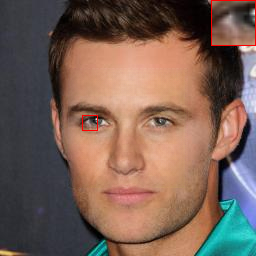}
	}
	
	\centering
	\subfloat{
		\centering
		\includegraphics[width=.13\textwidth]{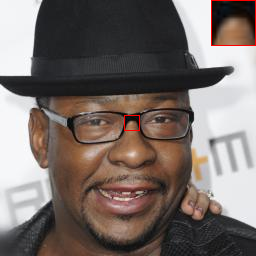}
	}
	\subfloat{
		\centering
		\includegraphics[width=.13\textwidth]{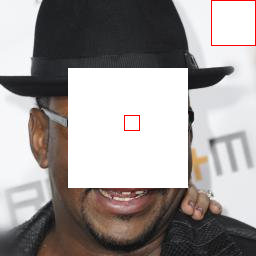}
	}
	\subfloat{
		\centering
		\includegraphics[width=.13\textwidth]{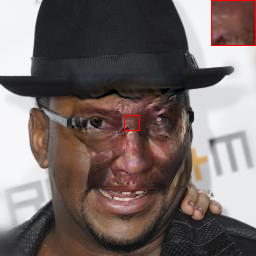}
	}
	\subfloat{
		\centering
		\includegraphics[width=.13\textwidth]{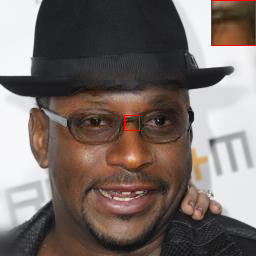}
	}
	\subfloat{
		\centering
		\includegraphics[width=.13\textwidth]{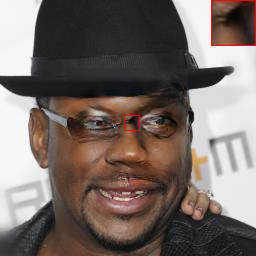}
	}
	\subfloat{
		\centering
		\includegraphics[width=.13\textwidth]{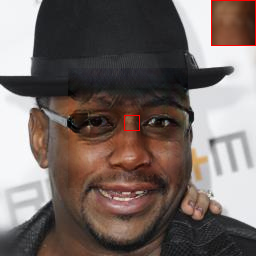}
	}
	\subfloat{
		\centering
		\includegraphics[width=.13\textwidth]{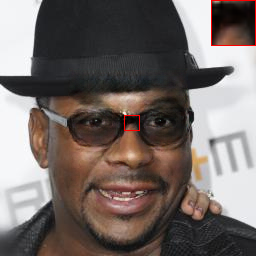}
	}
	
	\centering
	\subfloat{
		\centering
		\includegraphics[width=.13\textwidth]{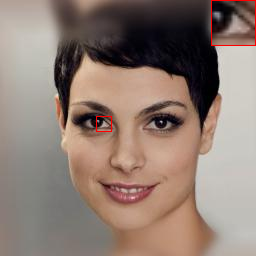}
	}
	\subfloat{
		\centering
		\includegraphics[width=.13\textwidth]{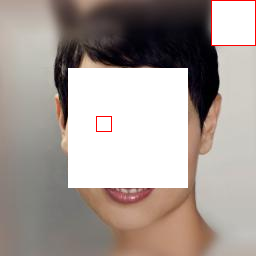}
	}
	\subfloat{
		\centering
		\includegraphics[width=.13\textwidth]{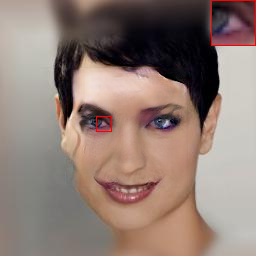}
	}
	\subfloat{
		\centering
		\includegraphics[width=.13\textwidth]{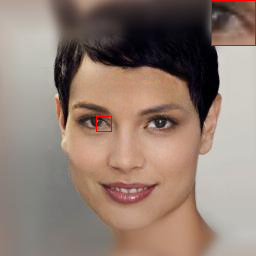}
	}
	\subfloat{
		\centering
		\includegraphics[width=.13\textwidth]{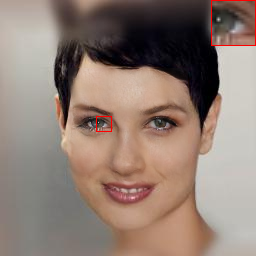}
	}
	\subfloat{
		\centering
		\includegraphics[width=.13\textwidth]{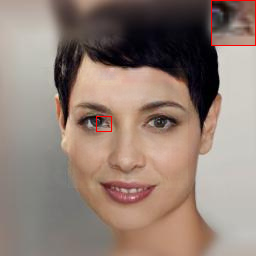}
	}
	\subfloat{
		\centering
		\includegraphics[width=.13\textwidth]{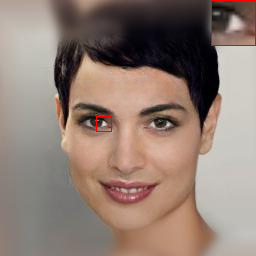}
	}
	
	\centering
	\subfloat{
		\centering
		\includegraphics[width=.13\textwidth]{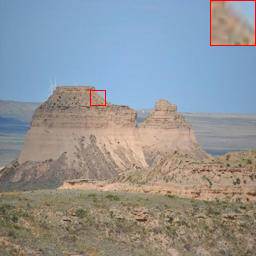}
	}
	\subfloat{
		\centering
		\includegraphics[width=.13\textwidth]{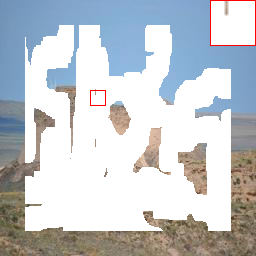}
	}
	\subfloat{
		\centering
		\includegraphics[width=.13\textwidth]{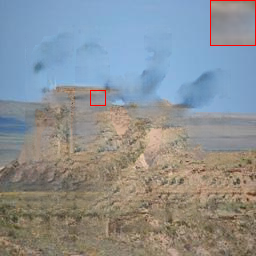}
	}
	\subfloat{
		\centering
		\includegraphics[width=.13\textwidth]{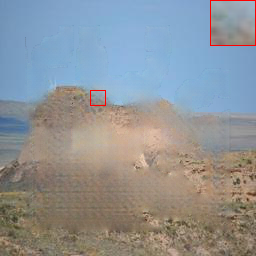}
	}
	\subfloat{
		\centering
		\includegraphics[width=.13\textwidth]{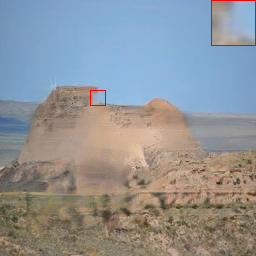}
	}
	\subfloat{
		\centering
		\includegraphics[width=.13\textwidth]{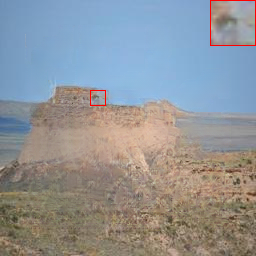}
	}
	\subfloat{
		\centering
		\includegraphics[width=.13\textwidth]{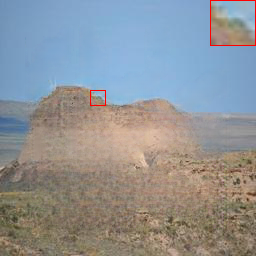}
	}
	
	\centering
	\subfloat{
		\centering
		\includegraphics[width=.13\textwidth]{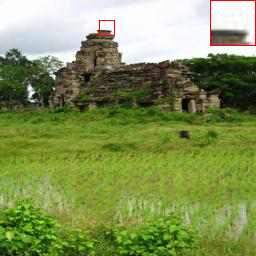}
	}
	\subfloat{
		\centering
		\includegraphics[width=.13\textwidth]{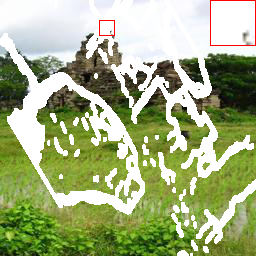}
	}
	\subfloat{
		\centering
		\includegraphics[width=.13\textwidth]{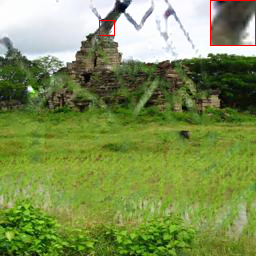}
	}
	\subfloat{
		\centering
		\includegraphics[width=.13\textwidth]{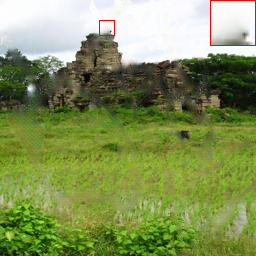}
	}
	\subfloat{
		\centering
		\includegraphics[width=.13\textwidth]{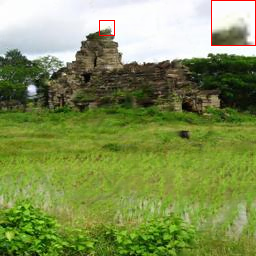}
	}
	\subfloat{
		\centering
		\includegraphics[width=.13\textwidth]{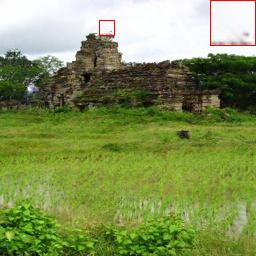}
	}
	\subfloat{
		\centering
		\includegraphics[width=.13\textwidth]{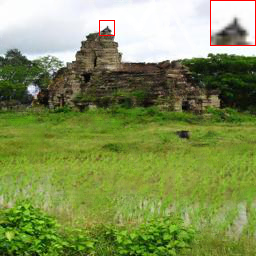}
	}
	
	\setcounter{subfigure}{0}
	\subfloat[GT.]{
		\centering
		\includegraphics[width=.13\textwidth]{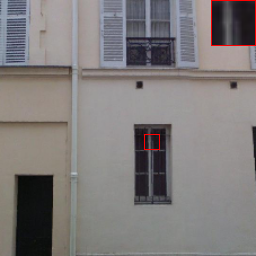}
	}
	\subfloat[Input]{
		\centering
		\includegraphics[width=.13\textwidth]{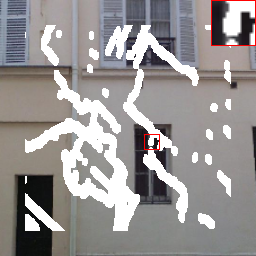}
	}
	\subfloat[CA]{
		\centering
		\includegraphics[width=.13\textwidth]{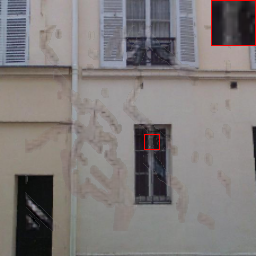}
	}
	\subfloat[SH]{
		\centering
		\includegraphics[width=.13\textwidth]{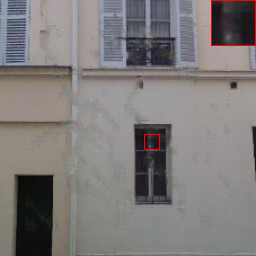}
	}
	\subfloat[GC]{
		\centering
		\includegraphics[width=.13\textwidth]{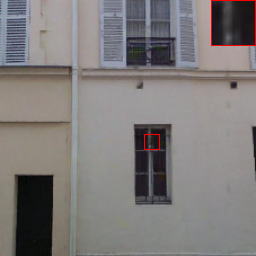}
	}
	\subfloat[EC]{
		\centering
		\includegraphics[width=.13\textwidth]{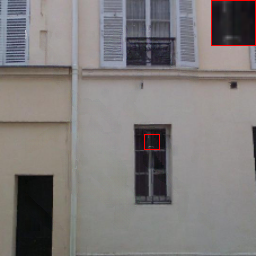}
	}
	\subfloat[Ours]{
		\centering
		\includegraphics[width=.13\textwidth]{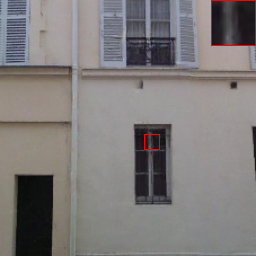}
	}
	\caption{Qualitative comparisons for image inpainting performance with centering and irregular masks on \texttt{CelebA-HQ}, \texttt{Places} and \texttt{Paris StreetView}. For each row, the images from left to right are ground truth, input images with centering or irregular mask, results generated by CA \cite{yu2018generative}, SH \cite{yan2018shift}, GC \cite{yu2019free}, EC \cite{nazeri2019edgeconnect} and our proposed method, respectively.}
	\label{fig:experimental_results}
\end{figure*}

\section{Experiments}\label{sec:experiments}

The proposed deep generative model is implemented using PyTorch framework. The training is performed on a desktop equipped with a Core-i7 and a single GTX 2080 GPU. To stabilize the training process and alleviate the gradient vanishing problem, we first train the generator $G_1$ and the discriminator $D_1$ in the LBP network. Then we concatenate the $G_1$ to the image inpainting network, and perform an end-to-end training over $G_1$, $G_2$ and $D_2$ simultaneously. Adam \cite{kingma2014adam} algorithm is adopted, where the parameters in Adam are $\beta_1=0.5$, $\beta_2=0.999$ and learning rate $r=2\times10^{-4}$. We train the model with the batch size of 1 and set the parameters of the spatial attention layer to be $h=13$ and $T=2$. In the loss functions, the parameters trading off different terms are set to be $\lambda_m=0.01$, $\lambda_r=10$, $\lambda_{a}=0.2$, $\lambda_p=1$ and $\lambda_s=10$. Note that all the parameters are fixed when performing the subsequent experiments.

We evaluate the inpainting performance of our method over three publicly available datasets: a high-quality human face dataset \texttt{CelebA-HQ} \cite{karras2017progressive}, a natural scene dataset collected from the real world \texttt{Places} \cite{zhou2017places}, and a street view dataset containing various objects \texttt{Paris StreetView} \cite{doersch2012makes}. The \texttt{CelebA-HQ} dataset contains 28000 training images and 2000 testing images. The \texttt{Places} dataset includes 365 categories, each containing 5000 training images and 100 validation images, where these validation images are used for testing. The \texttt{Paris StreetView} dataset has 14900 images in the training set and 100 images in the testing set.

For comparison purpose, we adopt four state-of-the-art inpainting methods: Contextual Attention (CA) \cite{yu2018generative}, Shift-net (SH) \cite{yan2018shift}, Gated Convolutions (GC) \cite{yu2019free}, and EdgeConnect (EC) \cite{nazeri2019edgeconnect}, with both the centering and the irregular loss patterns. Same with the settings of the aforementioned methods, the centering masks are of sizes $120 \times 120$, and irregular masks are obtained from \cite{liu2018image}, with various missing-to-known area ratios.

\subsection{Qualitative Comparisons}
Fig. \ref{fig:experimental_results} shows the inpainting results of different algorithms for some representative testing images. Additional inpainting results can be found in the complementary materials. For CA \cite{yu2018generative}, the main semantic information of the missing area can be well restored; but the incoherent artifacts around the boundary are quite obvious (e.g., the first three rows). This phenomenon is especially visible when the missing region is located in the homogenous parts (e.g., the sixth row). SH \cite{yan2018shift} can generate visually much more realistic images due to the shift layer, the guidance loss and the improved network architecture. However,the boundary inconsistency is still very severe (e.g., the fourth row), and often some fine textures are highly blurry (e.g., the fifth row). \setlength{\tabcolsep}{4pt}
\begin{table}[t]
	\caption{Quantitative comparisons over \texttt{CelebA-HQ} with centering mask among CA \cite{yu2018generative}, SH \cite{yan2018shift}, GC \cite{yu2019free}, EC \cite{nazeri2019edgeconnect} and ours. $^-$Lower is better. $^+$Higher is better}
	\label{tab:centering}
	\begin{center}
		\begin{tabular}{lccccc}
			\hline
			\noalign{\smallskip}
			Method & CA \cite{yu2018generative} & SH \cite{yan2018shift} & GC \cite{yu2019free} & EC \cite{nazeri2019edgeconnect} & Ours \\
			\noalign{\smallskip}
			\hline
			\noalign{\smallskip}
			$\ell_1^-$(\%) & 4.98 & 3.20 & 4.46 & 3.64 & \textbf{3.16} \\
			SSIM$^+$ & 0.882 & 0.924 & 0.897 & 0.912 & \textbf{0.926} \\
			PSNR$^+$ & 25.05 & 28.13 & 26.43 & 27.20 & \textbf{28.34} \\
			\hline
		\end{tabular}
	\end{center}
\end{table}\setlength{\tabcolsep}{1.4pt}GC \cite{yu2019free} generally can produce pretty good results; but also leads to rather inconsistent, blurry artifacts in the texture regions and also visible distortions around the mask boundary (e.g., the fourth row). Furthermore, even though EC \cite{nazeri2019edgeconnect} can produce visually good results by first complementing the edge contours, some broken or blurred edges can be observed (e.g., the fourth and fifth rows). Also, in some cases, the mask boundaries are still quite obvious (e.g., the sixth row). This may be due to the inadequate guidance offered by the edges. Compared with these methods, our proposed model can learn more reasonable semantic relevance and generate more realistic inpainting results (especially those fine structures and texture regions), primarily thanks to the rich amount of guiding information provided by LBP learning and the employment of the new spatial attention layer in the inpainting network.

\setlength{\tabcolsep}{4pt}
\begin{table}
	\caption{Quantitative comparisons over \texttt{Places} with irregular mask among CA \cite{yu2018generative}, SH \cite{yan2018shift}, GC \cite{yu2019free}, EC \cite{nazeri2019edgeconnect} and ours. $^-$Lower is better. $^+$Higher is better}
	\begin{center}
		\begin{tabular}{lcccccc}
			\hline
			\noalign{\smallskip}
			Method & Mask & CA \cite{yu2018generative} & SH \cite{yan2018shift} & GC \cite{yu2019free} & EC \cite{nazeri2019edgeconnect} & Ours \\
			\noalign{\smallskip}
			\hline
			\noalign{\smallskip}
			\multirow{4}{*}{$\ell_1^-$(\%)} & 10-20\% & 5.11 & 2.82 & 3.07 & 2.78 & \textbf{2.43} \\
			& 20-30\% & 9.05 & 5.03 & 5.45 & 4.95 & \textbf{4.47} \\
			& 30-40\% & 13.40 & 7.53 & 8.25 & 7.42 & \textbf{6.84} \\
			& 40-50\% & 17.61 & 10.43 & 11.57 & 10.33 & \textbf{9.65} \\
			\hline
			\noalign{\smallskip}
			\multirow{4}{*}{SSIM$^+$} & 10-20\% & 0.887 & 0.921 & 0.929 & 0.924 & \textbf{0.936} \\
			& 20-30\% & 0.805 & 0.860 & 0.869 & 0.868 & \textbf{0.880} \\
			& 30-40\% & 0.724 & 0.793 & 0.802 & 0.804 & \textbf{0.817} \\
			& 40-50\% & 0.642 & 0.719 & 0.726 & 0.732 & \textbf{0.745} \\
			\hline
			\noalign{\smallskip}
			\multirow{4}{*}{PSNR$^+$} & 10-20\% & 24.08 & 27.93 & 27.67 & 27.29 & \textbf{28.85} \\
			& 20-30\% & 20.99 & 24.94 & 24.24 & 24.82 & \textbf{25.59} \\
			& 30-40\% & 18.92 & 22.80 & 21.82 & 22.63 & \textbf{23.30} \\
			& 40-50\% & 17.56 & 21.11 & 19.90 & 20.91 & \textbf{21.48} \\
			\hline
		\end{tabular}
	\end{center}
	\label{tab:irregular}
\end{table}
\setlength{\tabcolsep}{1.4pt}

\subsection{Quantitative Comparisons}
In addition to the qualitative comparisons, we also compare different methods quantitatively for both centering and irregular masks, as shown in Tables \ref{tab:centering}-\ref{tab:irregular}. Here, we adopt the commonly used metrics, namely, $\ell_1$ loss, peak signal-to-noise ratio (PSNR) and structural similarity index (SSIM). It can be seen that our method consistently outperforms all the competing algorithms.

\begin{figure}
	\centering
	\subfloat{
		\centering
		\includegraphics[width=.45\textwidth]{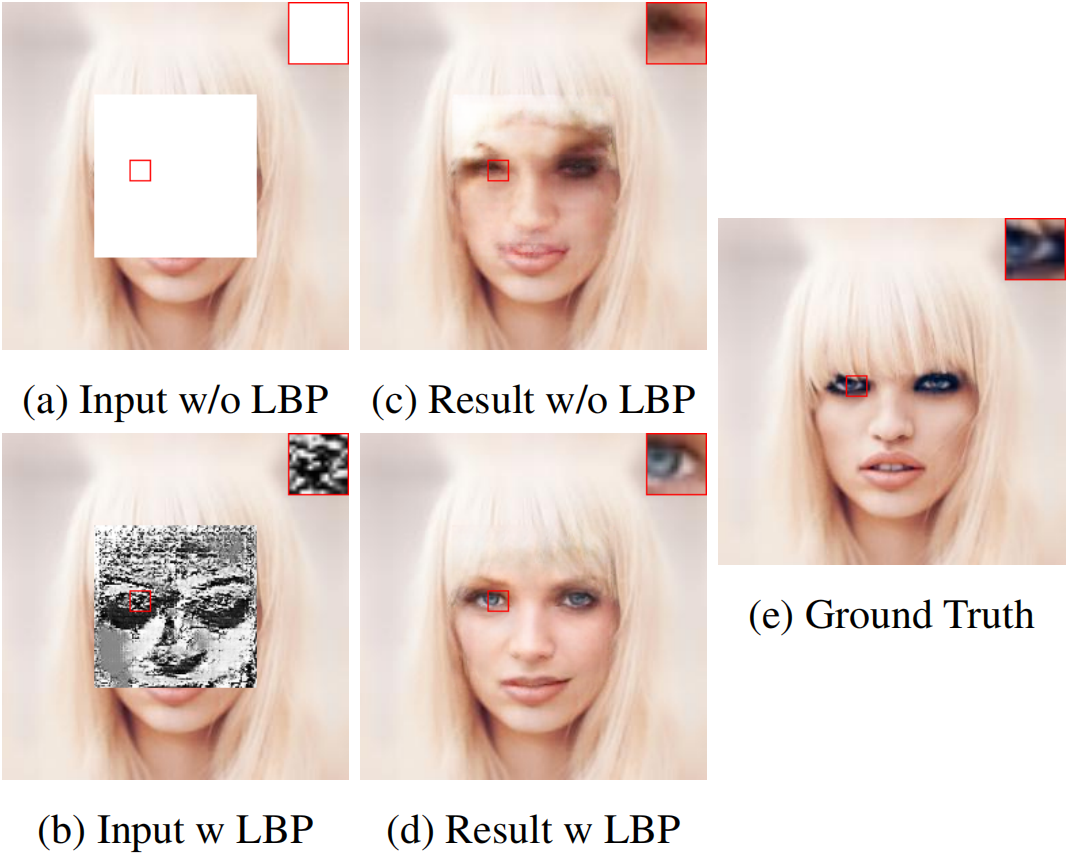}
	}
	\caption{Effect of the LBP learning network. (a)-(b) Inputs without or with the learned LBP. (c)-(d) Results without or with the LBP learning network. (e) Ground truth.}
	\label{fig:effect_lbp}
\end{figure}

\begin{figure}
	\centering
	\subfloat{
		\includegraphics[width=.11\textwidth]{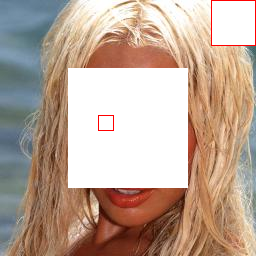}
	}
	\subfloat{
		\includegraphics[width=.11\textwidth]{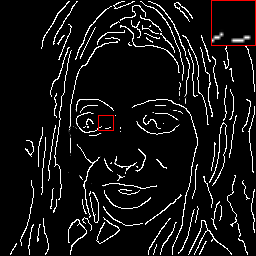}
	}
	\subfloat{
		\includegraphics[width=.11\textwidth]{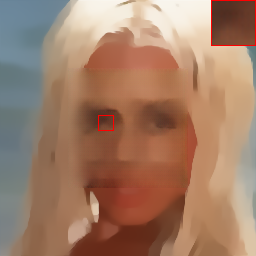}
	}
	\subfloat{
		\includegraphics[width=.11\textwidth]{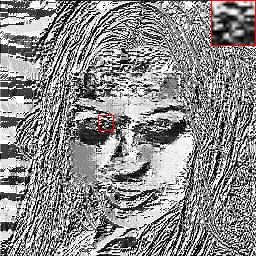}
	}
	\hfill
	\setcounter{subfigure}{0}
	\subfloat[]{
		\includegraphics[width=.11\textwidth]{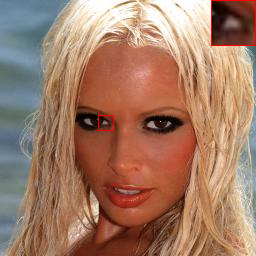}
	}
	\subfloat[]{
		\includegraphics[width=.11\textwidth]{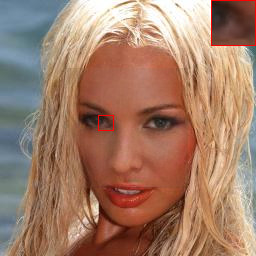}
	}
	\subfloat[]{
		\includegraphics[width=.11\textwidth]{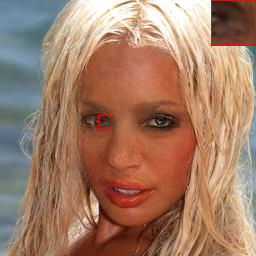}
	}
	\subfloat[]{
		\includegraphics[width=.11\textwidth]{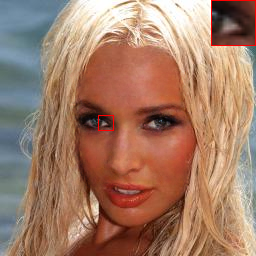}
	}
	\caption{Effect of the guidance provided by edges \cite{canny1986computational}, RTV \cite{xu2012RTV} and LBP \cite{ojala1996comparative}. (a) Input and ground truth. (b)-(d) The first row shows the predicted edges, RTV and LBP, and the second row presents corresponding results guided by them.}
	\label{fig:effect_edge_lbp}
\end{figure}

\subsection{Ablation Studies}
In this section, additional experiments are conducted to analyze how each component (e.g., LBP learning, the new spatial attention layer and the multi-level loss) of our proposed model contributes to the final inpainting results.

\subsubsection{Effect of the LBP learning network} To investigate the effectiveness of the LBP learning network, we inpaint the images with or without the learned LBP in the missing region, respectively. Fig. \ref{fig:effect_lbp} reports the comparison results, where Fig. \ref{fig:effect_lbp}(c) is the result without LBP learning, and Fig. \ref{fig:effect_lbp}(d) is the one guided by the learned LBP. It can be seen that, by learning the LBP of the missing region first, more structural information can be provided to significantly improve the inpainting performance, making the results sharper and more semantically reasonable.

Additionally, by replacing the LBP learning network with the edge generator \cite{nazeri2019edgeconnect} and structure reconstructor based on relative total variation (RTV) \cite{ren2019structureflow, xu2012RTV}, respectively, we explore the guidance provided by different structural information (edges, RTV or LBP) for performing the inpainting tasks. In Fig. \ref{fig:effect_edge_lbp}, we can see that the predicted edges contain discontinuities, and RTV, as an edge-preserved smooth method, inherently misses a lot of structural details. As a comparison, the sufficient structural information contained in LBP makes the result sharper (e.g., eyes and nose). This would suggest that LBP is a more appropriate candidate for providing structural information in the case of image inpainting.

\subsubsection{Effect of the spatial attention layer} 
\begin{figure}
	\subfloat[Input]{
		\includegraphics[width=.15\textwidth]{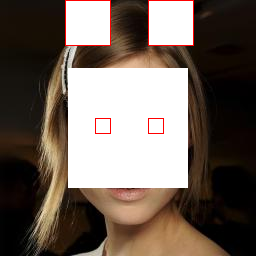}
	}
	\subfloat[w/o]{
		\includegraphics[width=.15\textwidth]{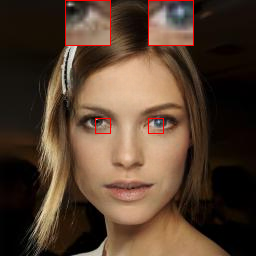}
	}
	\subfloat[w CA's \cite{yu2018generative} layer]{
		\includegraphics[width=.15\textwidth]{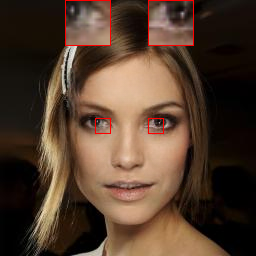}
	}
	\\
	\subfloat[w SH's \cite{yan2018shift} layer]{
		\includegraphics[width=.15\textwidth]{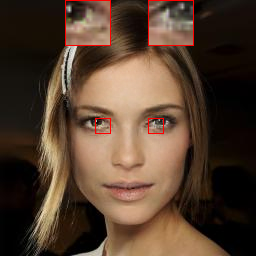}
	}
	\subfloat[w our layer]{
		\includegraphics[width=.15\textwidth]{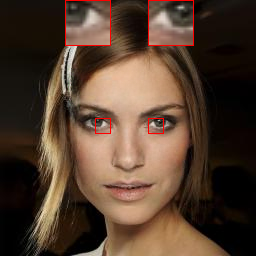}
	}
	\subfloat[Ground Truth]{
		\includegraphics[width=.15\textwidth]{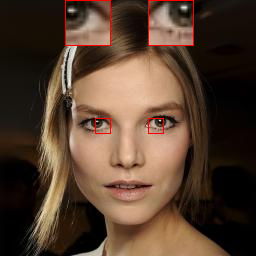}
	}
	\caption{Effect of different spatial attention layer. (a) Input. (b)-(e) Results without spatial layer, with CA's \cite{yu2018generative} layer, with SH's \cite{yan2018shift} layer and with our layer, respectively. (f) Ground truth.}
	\label{fig:effect_layer}
\end{figure}

We evaluate the effectiveness of our spatial attention layer by removing it, replacing it with the contextual attention layer \cite{yu2018generative} or the shift layer \cite{yan2018shift}, respectively. The results in Fig. \ref{fig:effect_layer} indicate that our spatial attention layer can better guarantee semantic coherence within the missing region (e.g., the restored eyes), while some artifacts and inconsistent contents are produced by CA's and SH's layer. The statistics in Table \ref{tab:effect_attention_layer} also verify the superiority of our proposed spatial attention layer quantitatively (e.g., over 0.3 dB PSNR gains).

\begin{table}
	\caption{Effect of different spatial attentions on \texttt{Places} with centering mask. $^-$Lower is better. $^+$Higher is better}
	\begin{center}
		\begin{tabular}{lcccc}
			\hline
			\noalign{\smallskip}
			Method ~~~& w/o ~~& w CA's ~~& w SH's ~~& w our \\
			\noalign{\smallskip}
			\hline
			\noalign{\smallskip}
			$\ell_1^-$(\%) & 7.68 & 7.64 & 7.65 & \textbf{7.33}\\
			SSIM$^+$ & 0.722 & 0.723 & 0.721 & \textbf{0.729} \\
			PSNR$^+$ & 22.33 & 22.39 & 22.41 & \textbf{22.72} \\
			\hline
		\end{tabular}
	\end{center}
	\label{tab:effect_attention_layer}
\end{table}

\subsubsection{Hyperparameters of spatial attention layer} The considerable operations of convolutional filters performed in the spatial attention layer may cause memory overhead for GPUs \cite{yu2018generative}. One of the main factors influencing the calculation time is the size of the feature map, i.e., when the index $h$ of the spatial attention layer is bigger, the feature map size goes larger, which requires more computational resources. However, when $h$ is smaller, the performance could be degraded due to the insufficient number of extracted patches. To achieve better tradeoff between the efficiency and the performance, we set the resolution of the attention layer to $32 \times 32$, i.e., $h$ = 13. Another influencing factor is the number of extracted patches used for spatial attention operations. Since negligible improvements can be brought when involving more patches to update the missing region, we heuristically determine $T=2$ to save the cost.

\subsubsection{Effect of the multi-level loss $\mathcal{L}_m$} 
\begin{figure}
	\subfloat{
		\includegraphics[width=.11\textwidth]{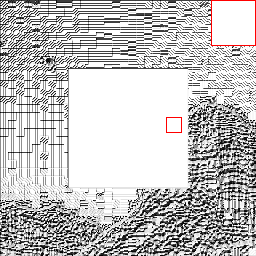}
	}
	\subfloat{
		\includegraphics[width=.11\textwidth]{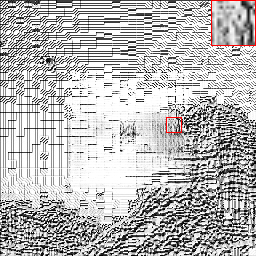}
	}
	\subfloat{
		\includegraphics[width=.11\textwidth]{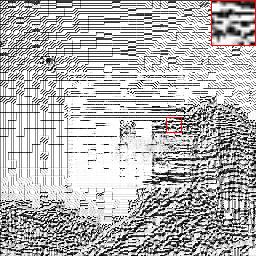}
	}
	\subfloat{
		\includegraphics[width=.11\textwidth]{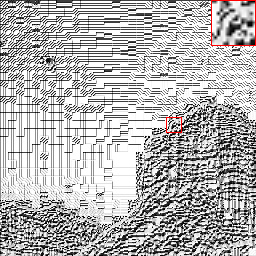}
	}
	\\
	\subfloat[Input]{
		\includegraphics[width=.11\textwidth]{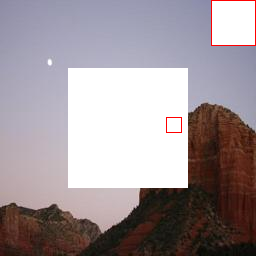}
	}
	\subfloat[w/o $\mathcal{L}_m$]{
		\includegraphics[width=.11\textwidth]{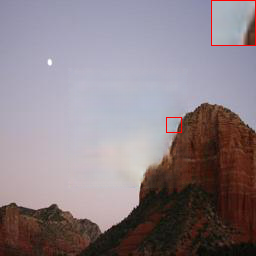}
	}
	\subfloat[w $\mathcal{L}_m$]{
		\includegraphics[width=.11\textwidth]{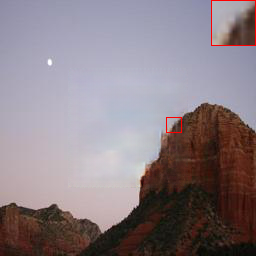}
	}
	\subfloat[GT.]{
		\includegraphics[width=.11\textwidth]{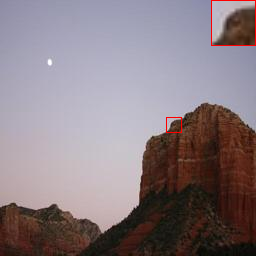}
	}
	\caption{Effect of the multi-level loss $\mathcal{L}_m$ on \texttt{Places} with a centering mask. (a) Inputs. (b)-(c) LBP and inpainting results of our model without or with $\mathcal{L}_m$. (d) Ground truth.}
	\label{fig:effect_multi_level}
\end{figure}

We evaluate the effectiveness of the multi-level loss by adding or dropping $\mathcal{L}_m$ in the loss functions of the LBP learning and the image inpainting networks. As shown in Fig. \ref{fig:effect_multi_level}, without the multi-level loss, the learned LBP could not satisfactorily recover some structural information, leading to inferior performance of the inpainting result. By incorporating the multi-level loss term, the LBP feature can be more faithfully predicted, and the inpainting result gets improved. Besides, quantitative comparisons in Table \ref{tab:effect_multi_level} also verify the superiority of multi-level loss (e.g., around 0.11 dB PSNR gains).

\begin{table}
	\caption{Effect of the multi-level loss $\mathcal{L}_m$ on \texttt{Places} with centering masks. $^-$Lower is better. $^+$Higher is better}
	\begin{center}
		\begin{tabular}{lccc}
			\hline
			\noalign{\smallskip}
			Method~~~ & $\ell_1^-$(\%)~ & SSIM$^+$~ & PSNR$^+$~ \\
			\noalign{\smallskip}
			\hline
			\noalign{\smallskip}
			w/o $\mathcal{L}_m$& 4.80 & 0.844 & 24.20 \\
			w $\mathcal{L}_m$ & \textbf{4.71} & \textbf{0.845} & \textbf{24.31} \\
			\hline
		\end{tabular}
	\end{center}
	\label{tab:effect_multi_level}
\end{table}

\section{Conclusions}\label{sec:conclusions}
In this paper, we have proposed a deep generative model for image inpainting with LBP learning and a new spatial attention mechanism. The proposed model has been formed with two networks: a LBP learning network, which aims to learn the LBP feature of the missing region, and an image inpainting network, which generates the inpainting results by using the learned LBP as a guidance. Within the two-stage framework, how to select these two sub-networks is very crucial and could lead to vastly different results. Compared with edges and RTV, LBP features contain much more structural information and are almost parameter-free. Furthermore, we have designed a new spatial attention layer, and have incorporated it into the image inpainting network. The proposed spatial attention strategy not only considers the dependency between the know region and the filled region, but also the one within the filled region. Such dependency, though obviously important, has been overlooked by all the existing schemes. Experimental results have been provided to demonstrate the superiority of the proposed model.


%

\appendices
\section{}
The architectures of the generators $G_1$ or $G_2$, the discriminators $D_1$ or $D_2$ are shown in Table \ref{tab:g1g2}. Conv($f, k, s, p$) means a convolutional layer with $f$ filters, kernel size $k$, stride $s$ and padding $p$. DeConv denotes deconvolutional layer. IN represents InstanceNorm and LReLU is LeakyReLU with slop of $0.2$. Cat(Layer $b_1$, Layer $b_2$) concatenates the outputs of Layer $b_1$ and Layer $b_2$. Tanh and Sigmoid are the activation functions. We embed our spatial attention layer \textit{only} in Layer 13 of $G_2$.

\begin{table}[h]
	\caption{The architectures of the generators $G_1$ or $G_2$, and the discriminators $D_1$ or $D_2$.}
	\label{tab:g1g2}
	\begin{center}
		\begin{tabular}{|l|}
			\hline
			The architecture of the generators $G_1$ or $G_2$ \\
			\hline
			[Layer 1] Conv(64, 4, 2, 1);\\
			\hline
			[Layer 2] LReLU; Conv(128, 4, 2, 1); IN;\\
			\hline
			[Layer 3] LReLU; Conv(256, 4, 2, 1); IN;\\
			\hline
			[Layer 4] LReLU; Conv(512, 4, 2, 1); IN;\\
			\hline
			[Layer 5] LReLU; Conv(512, 4, 2, 1); IN;\\
			\hline
			[Layer 6] LReLU; Conv(512, 4, 2, 1); IN;\\
			\hline
			[Layer 7] LReLU; Conv(512, 4, 2, 1); IN;\\
			\hline
			[Layer 8] LReLU; Conv(512, 4, 2, 1);
			ReLU; DeConv(512, 4, 2, 1); IN;\\
			\hline
			[Layer 9] Cat(Layer 8, Layer 7);
			ReLU; DeConv(512, 4, 2, 1); IN;\\
			\hline
			[Layer 10] Cat(Layer 9, Layer 6);
			ReLU; DeConv(512, 4, 2, 1); IN;\\
			\hline
			[Layer 11] Cat(Layer 10, Layer 5);
			ReLU; DeConv(512, 4, 2, 1); IN;\\
			\hline
			[Layer 12] Cat(Layer 11, Layer 4);
			ReLU; DeConv(256, 4, 2, 1); IN;\\
			\hline
			[Layer 13] Cat(Layer 12, Layer 3);\\
			~~~~~~~~~~~~~~\textbf{(SpatialAttention)};\\
			~~~~~~~~~~~~~~ReLU; DeConv(128, 4, 2, 1); IN;\\
			\hline
			[Layer 14] Cat(Layer 13, Layer 2);
			ReLU; DeConv(64, 4, 2, 1); IN;\\
			\hline
			[Layer 15] Cat(Layer 14, Layer 1);
			ReLU; DeConv(3, 4, 2, 1); Tanh;\\
			\hline
		\end{tabular}
	\end{center}
	
	\begin{center}
		\begin{tabular}{|l|}
			\hline
			The architectures of the discriminators $D_1$ or $D_2$ \\
			\hline
			[Layer 1] Conv(64, 4, 2, 1);\\
			\hline
			[Layer 2] LReLU; Conv(128, 4, 2, 1); IN;\\
			\hline
			[Layer 3] LReLU; Conv(256, 4, 2, 1); IN;\\
			\hline
			[Layer 4] LReLU; Conv(512, 4, 2, 1); IN;\\
			\hline
			[Layer 5] LReLU; Conv(1, 4, 2, 1); Sigmoid;\\
			\hline
		\end{tabular}
	\end{center}
\end{table}



\ifCLASSOPTIONcaptionsoff
  \newpage
\fi

\bibliographystyle{IEEEtran}
\bibliography{ref}

\end{document}